\documentclass[letterpaper, 10 pt, conference]{ieeeconf}  

\IEEEoverridecommandlockouts                              

\usepackage{graphicx}
\usepackage{amsmath} 
\usepackage{hyperref}

\title{\LARGE \bf
To What Extent Does the Perceived Obesity Level of \\ Humanoid Robots Affect People's Trust in Them?}

\author{Yoav Yoscovich$^{1}$, Amir Schreiber$^{1}$, Nir Hadar$^{1}$, Reuth Mirsky$^{1,2}$
\thanks{$^{1}$Department of Computer Science, Bar Ilan University}%
\thanks{$^{2}$Department of Computer Science, Tufts University       {\tt\small reuth.mirsky@tufts.edu}}%
}

\begin{document}

\maketitle
\thispagestyle{empty}
\pagestyle{empty}

\begin{abstract}

Despite obesity being widely discussed in the social sciences, the effect of a robot's perceived obesity level on trust is not covered by the field of HRI. While in research regarding humans, \textit{Body Mass Index (BMI)} is commonly used as an indicator of obesity, this scale is completely irrelevant in the context of robots, so it is challenging to operationalize the perceived obesity level of robots; indeed, while the effect of robot's size (or height) on people's trust in it was addressed in previous HRI papers, the perceived obesity level factor has not been addressed. This work examines to what extent the perceived obesity level of humanoid robots affects people's trust in them. To test this hypothesis, we conducted a within-subjects study where, using an online pre-validated questionnaire, the subjects were asked questions while being presented with two pictures of humanoids, one with a regular obesity level and the other with a high obesity level. The results show that humanoid robots with lower perceived obesity levels are significantly more likely to be trusted.

\end{abstract}

\section{Introduction}
A broad spectrum of literature has developed, emphasizing the importance of human-robot trust \cite{trustInHRIRevisited}. More importantly, studies suggest that the more important elements of trust are robot-related (e.g., robot attributes), namely, things that robot designers can directly manipulate \cite{metaAnalysisHRL}. Therefore, it is crucial to understand how simple changes in a robot's design can improve, or at least not thwart, the potential for trust between humans and robots in their interaction. \textbf{This paper presents an initial investigation of how a robot's perceived obesity will affect people's trust.}
Research in social sciences has tested the effect of physicians' body weight on patients' level of trust. For example, it has been found that health providers’ excess weight may negatively affect patients’ perceptions of their credibility, level of trust, and inclination to follow medical advice \cite{obeseDoctorsAffectTrust}. It should be noted that findings suggest the difference in trust is not necessarily related to medical counseling specifically regarding weight, but a general unwillingness of patients to follow the medical advice of overweight doctors, regardless of the medical field the medical advice is related to \cite{obeseDoctorsAffectTrust2}.
In this project, we aim to identify whether the same conclusions can be drawn in the context of robots; in other words, does \textit{weightism} (bias or discrimination against people who are overweight) exist not only towards people but also towards humanoid robots? It should be noted that the effect of a robot's size on people's trust in it was addressed in previous HRI papers; various articles have researched the effects of a robot's \textit{height} on how people perceive and feel about the robot \cite{heightEffect1}. However, many articles have used the terms "size" and "height" interchangeably when referring to the dimensions of the robot \cite{heightEffect2}. In contrast, we wanted to study the effect of the perceived obesity of the robot.

\section{Experimental Design}
In this study, we used a Google Forms online questionnaire\footnote{A link to the Google Forms questionnaire sent to the subjects: \url{https://forms.gle/Ayh8APWoyTyL2KZU8}.}. Each trial involves one subject and does not require any confederates or experimenters to monitor or instruct the participants during the trial. Namely, as detailed below, all the relevant information is presented to the participants as part of the online questionnaire. 

\begin{figure}
    \centering
    \includegraphics[width=0.8\linewidth]{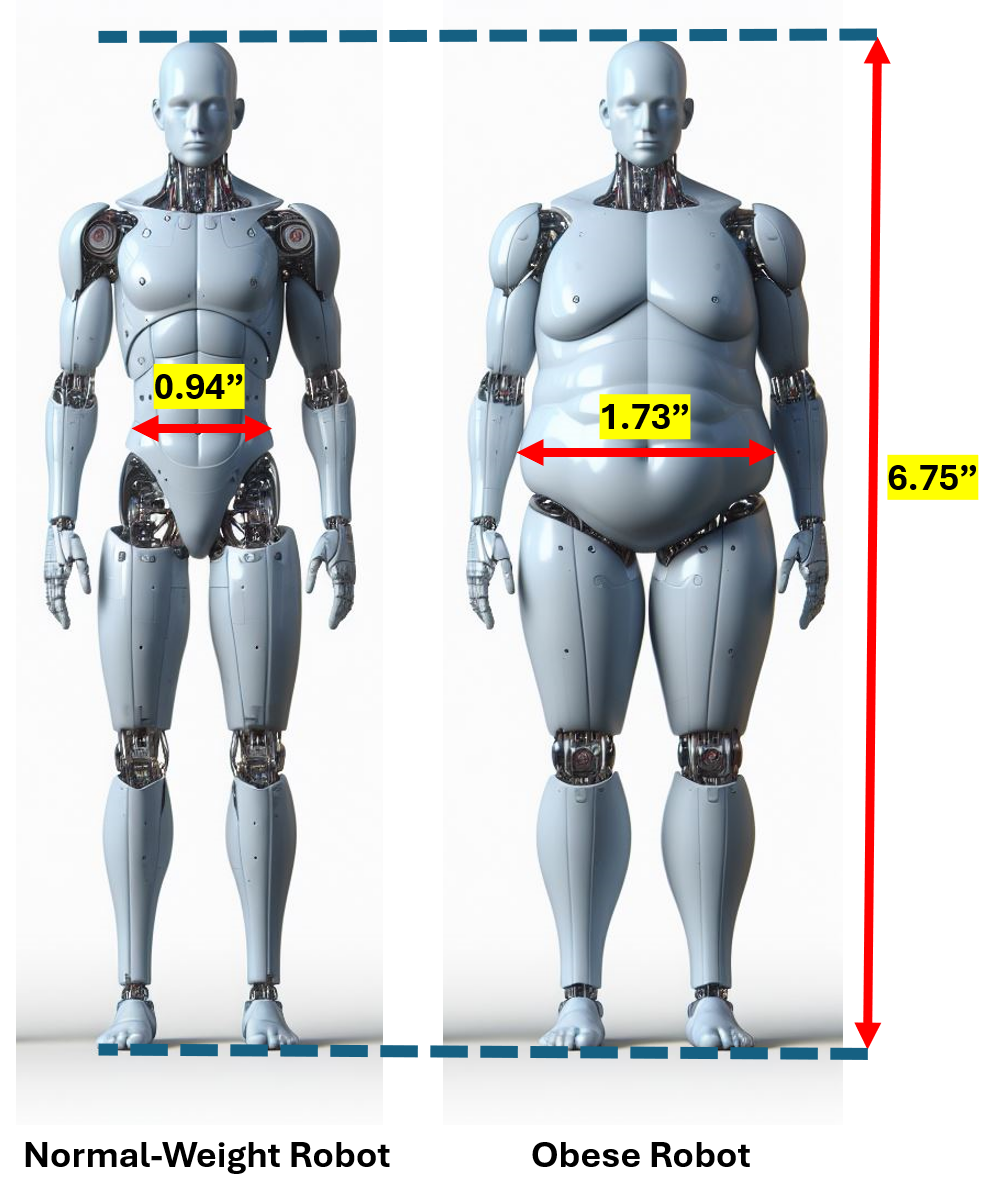}
    \caption{Example of different variations of a robot as used in this experiment, showing the difference in height-weight ratio between the normal-weight robot and the obese robot.}
    \label{fig:enter-label}
\end{figure}

    This is a \textit{within-subjects} experimental design; namely, each participant has experienced the two conditions - two robots with two different perceived weight levels: a \textit{normal-weight} robot and an \textit{obese} robot, such that given the same height, the ratio between the robot's circumference is 1.84 in favor of the obese-looking robot (see Figure 1 for details). We compared the different experiences of each participant. 
    We have randomized the order of the conditions that the subjects experience. 
    
    
    The different \textit{conditions} presented to the subjects are two pictures of robots: one of them looks normal-weight, and the other looks obese. Many efforts have been made to ensure that this distinction is not only made by eyesight but according to mathematical calculations and an official definition for ``obesity'' \cite{apovian2016obesity}. We have used \textit{WHtR} (Waist-to-Height Ratio) to operationalize the construct of \textit{perceived obesity level of the robot}. We have followed official guidelines to ensure that the two conditions sufficiently differ one from another; Health agencies suggest that a \textbf{WHtR cutoff of \textit{0.5}} can be used in different sex and ethnic groups and is generally accepted as a universal cutoff for central obesity in children (aged $\geq$ 6 years) and adults \cite{0.5Threshold, ashwell2009obesity}. 
    %
    %
    According to the calculation presented there, we got the following ratios: 
    \begin{equation*}
    \text{WHtR}_{\text{normal-robot}} = \frac{\text{WC}_{\text{normal-robot}}}{H_{\text{normal-robot}}} = \frac{2 \times 0.94"}{6.75"} = \mathbf{0.279 < 0.5}  
    \end{equation*}

     \begin{equation*}
    \text{WHtR}_{\text{obese-robot}} = \frac{\text{WC}_{\text{obese-robot}}}{H_{\text{obese-robot}}} = \frac{2 \times 1.73"}{6.75"} = \mathbf{0.513 > 0.5}     
     \end{equation*}
    
    It should be noted that the actual WHtR of the robots is likely to be slightly larger than our calculation since, due to 2D limitations, we have neglected the size of the robot's side of the body. 
    Additionally, control variables that were consistent in both conditions are the humanoid's height, facial expressions, and overall features (except for the robot's perceived obesity level). Namely, we have made efforts to make sure that the \underline{only} difference between the two robots used in our experiment is their perceived obesity level. In particular, various articles have researched the effects of robot's \textit{height} on how people perceive and feel about the robot \cite{heightEffect1,heightEffect2}; therefore, we have made efforts to ensure that this factor indeed remains consistent between the conditions, and indeed, the height of both humanoids used is equal. We verified that in two ways while testing the Google Forms questionnaire: First, we verified that the WHtR values of each robot remain consistent among all screens aspect ratios. Second, that each subject answering the questionnaire is presented with two robots of the same height. Thus, even if the screen size of the device used is small, the pictures of the two robots will be smaller respectively.
    To generate the images of the robot, We reviewed several generative AI options, including \textit{Midjourney}, but after trying different prompts in different generative AI services, we got the best results (most distinct in perceived weight, yet identical in all their other characteristics) using OpenAI's DALL-E 3\footnote{\url{https://openai.com/index/dall-e-3/}}. 
    The questionnaire used in this experiment consists of 40 questions per each condition (80 questions in total), taken from a pre-validated scale for humans' trust in robots, particularly relating to trust perception in HRI \cite{validatedTrustScaleInHRI}. 
    Each question is answered using an 11-point \textit{likert scale} (0\%, 10\%, ..., 100\%). It should be noted that to keep the subjects vigilant, a high trust level was either related to answering 0\% or 100\% (some questions are \textit{reverse-coded}). 
%
    In addition, at the beginning of the questionnaire, we ask the participants to provide their \textit{gender} (the options are: ``male'', ``female'' or ``prefer not to say''), and to choose their appropriate \textit{age range} (18-24, 25-34, 35-44, 45-54, 55-64, 65-74, 75-84, 85+). 

\section{Results}

We had 20 participants as subjects in the study, 60\% of them were male and 40\% female (none of the participants answered ``prefer not to say''). Half of the participants were at the 18-34 age groups, and one participant was in the 85+ age group.
The average total score across all participants is 29.87 out of a range [0-40] for the normal-looking robot and 20.59 for the obese-looking robot. The average trust score for each robot (which can also be calculated by dividing the average total score by 40) is 
%
%
$74.67\%$ ($stdv=0.15$) for the normal-looking robot, while the average trust score for the obese looking robot is $51.47\%$ ($stdv=0.14$). Overall, participants demonstrated a significantly higher trust in the normal-looking robot than in the obese looking one, $t(19)=5.54, p<.001$ (one-tailed).

\section{Conclusion and Future Work}
This project's purpose is to test how the perceived obesity level of humanoid robots affects people’s trust in them. Our preliminary results suggest that the perceived obesity level significantly influences trust levels. Specifically, participants demonstrated higher trust in the normal-looking robot, with an average trust score of 74.67\%, compared to the obese-looking robot, which received an average trust score of 51.47\%. This highlights the role of robots' appearance in human-robot interactions, emphasizing the importance of designing humanoid robots that are aesthetically appealing and aligned with societal expectations.
However, one should bear in mind that the obesity level may impact user trust differently depending on the robot's overall appearance and the specific interaction. Even when the only difference between two robots is their perceived obesity level, the difference in trust may vary for different pairs of robots with different designs. 
It is also interesting to investigate how people perceive underweight robots and whether underweight has similar effects on trust.

Additionally, obesity can be designed in various ways, and some designs may make the robot look less aesthetic, reducing its acceptance and trust. Hence, further research could test whether the same results are observed for different robots and obesity designs.
It should also be noted that different cultures may perceive obesity and other physical attributes differently, impacting the acceptance of humanoid robots. For example, it has been found that in many African countries, being overweight has been associated with richness, health, strength, and fertility \cite{africa}. Therefore, future research could explore these cultural dynamics by designing humanoid robots with varying obesity levels and measuring trust in these robots across diverse populations. This approach could lead to more effective robot designs tailored to the cultural preferences and societal norms of the target populations.

\bibliography{bibliography}
\bibliographystyle{unsrt}

\end{document}